\pdfoutput=1

\documentclass[11pt]{article}

\usepackage[]{acl}

\usepackage{times}
\usepackage{latexsym}

\usepackage[utf8]{inputenc} 
\usepackage[T1]{fontenc}    
\usepackage{hyperref}       
\usepackage{cleveref}
\usepackage{url}            
\usepackage{booktabs}       
\usepackage{amsfonts}       
\usepackage{nicefrac}       
\usepackage{microtype}      
\usepackage{xcolor}         
\usepackage{listings}
\usepackage{algorithm}
\usepackage{algorithmic}
\usepackage{multirow}
\usepackage{graphicx}
\usepackage{multicol}
\usepackage{CJKutf8}
\usepackage{booktabs}
\usepackage{subcaption}
\usepackage[T1]{fontenc}
\usepackage[utf8]{inputenc}
\usepackage{titlesec}
\usepackage[normalem]{ulem}
\usepackage{array}
\usepackage{amssymb}
\usepackage{wrapfig,lipsum,booktabs}
\usepackage{multirow}
\usepackage{graphicx}
\usepackage{subcaption}
\usepackage{pythonhighlight}
\usepackage{grffile}
\usepackage{bbm}
\usepackage[T1]{fontenc}

\usepackage[utf8]{inputenc}

\usepackage{microtype}

\usepackage{inconsolata}

\title{AdaMergeX: Cross-Lingual Transfer with Large Language Models via Adaptive Adapter Merging}

\author{%
  Yiran Zhao$^{1,2}$\footnotemark[1]\thanks{\;\;This work was done during the internship of Yiran Zhao and Huiming Wang at Alibaba DAMO Academy.} \quad Wenxuan Zhang$^{1,3}$\footnotemark[2]\thanks{\;\;Wenxuan Zhang is the corresponding author.} \quad Huiming Wang$^{1,4}$\footnotemark[1] \quad Kenji Kawaguchi$^{2}$
   \quad Lidong Bing$^{1,3}$ \\
   $^1$ DAMO Academy, Alibaba Group, Singapore \quad  $^2$ National University of Singapore \\ 
  $^3$ Hupan Lab, 310023, Hangzhou, China \quad $^4$ Singapore University of Technology and Design \\
   \texttt{zhaoyiran@u.nus.edu\quad kenji@comp.nus.edu.sg} \\
   \texttt{\{saike.zwx, huiming.wang, l.bing\}@alibaba-inc.com}
}

\begin{document}
\maketitle
\begin{abstract}

As an effective alternative to the direct fine-tuning on target tasks in specific languages, cross-lingual transfer addresses the challenges of limited training data by decoupling ``task ability'' and ``language ability'' by fine-tuning on the target task in the source language and another selected task in the target language, respectively. However, they fail to fully separate the task ability from the source language or the language ability from the chosen task. In this paper, we acknowledge the mutual reliance between task ability and language ability and direct our attention toward the gap between the target language and the source language on tasks. As the gap removes the impact of tasks, we assume that it remains consistent across tasks. Based on this assumption, we propose a new cross-lingual transfer method called \texttt{AdaMergeX} that utilizes adaptive adapter merging. By introducing a reference task, we can determine that the divergence of adapters fine-tuned on the reference task in both languages follows the same distribution as the divergence of adapters fine-tuned on the target task in both languages. Hence, we can obtain target adapters by combining the other three adapters. Furthermore, we propose a structure-adaptive adapter merging method. Our empirical results demonstrate that our approach yields new and effective cross-lingual transfer, outperforming existing methods across all settings.\footnote{Our code implementation is publicly available at \href{https://github.com/DAMO-NLP-SG/AdaMergeX}{https://github.com/DAMO-NLP-SG/AdaMergeX}}
\end{abstract}

\section{Introduction}

Multilingual NLP models, including conventional models such as mBERT~\citep{kenton2019bert}, XLM~\citep{conneau2019cross}, XLM-R~\citep{conneau2020unsupervised}, as well as recent multilingual large language models (LLMs) like ChatGPT~\citep{chatgpt}, PaLM2~\citep{anil2023palm}, Llama2~\citep{touvron2023llama}, have gained significant attention given the growing need for multilingual requirements. To further enhance the model's multilingual capability, particularly in cases where training data of certain tasks for low-resource languages is scarce and fine-tuning becomes impractical~\citep{ma2023prompt}, cross-lingual transfer is introduced to extend the task-solving ability in a source language to a wide range of target languages~\citep{lin2019choosing, chen2022towards, deb2023zero}. 

Essentially, cross-lingual transfer aims to transfer the ability to solve a certain task (``task-ability'') from a source language to a particular target language (``language ability''). 
Some cross-lingual transfer techniques do not directly improve the language ability in specific languages. Instead, they utilize the language ability in English for multilingual tasks by employing methods such as translation~\citep{liang2023machine, huang2023not}, representation alignment~\citep{nguyen2023enhancing, salesky2023pixel, gao2023improving}, or prompting method specifically developed for LLMs~\citep{li2023enhancing, tanwar2023multilingual, m3exam}. Some works intertwine these two abilities and utilize translated parallel corpora for fine-tuning~\citep{pan2021contrastive, zhang2022contrastive, zhu2023extrapolating}.

\begin{figure*}[t]
\centering
\includegraphics[width=0.38\textwidth]{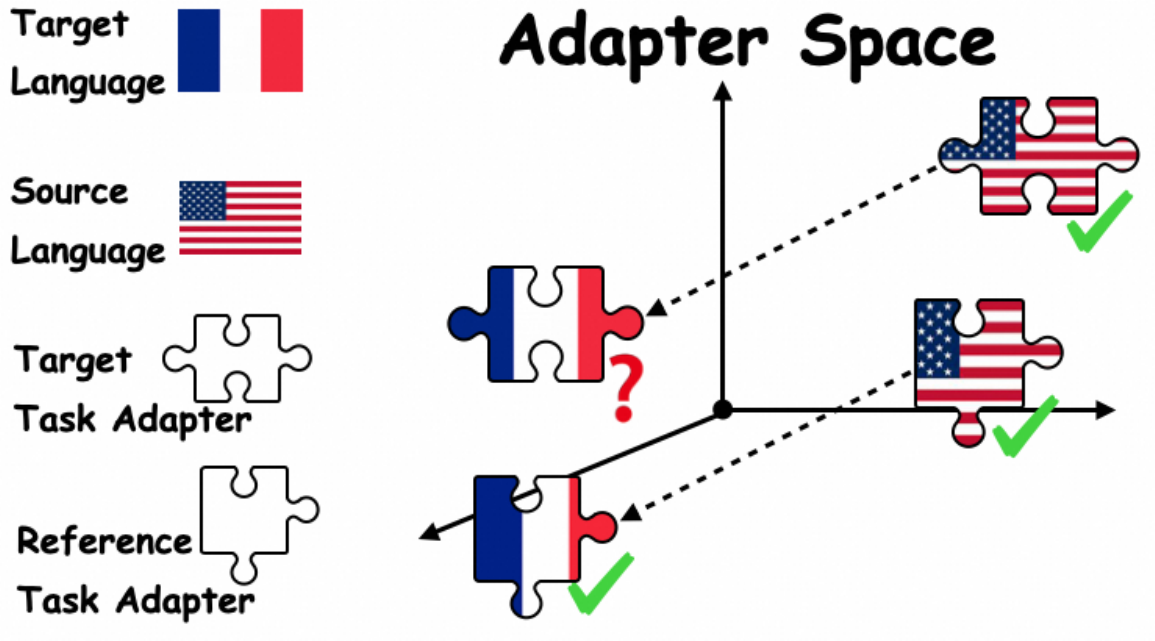}
\includegraphics[width=0.6\textwidth]{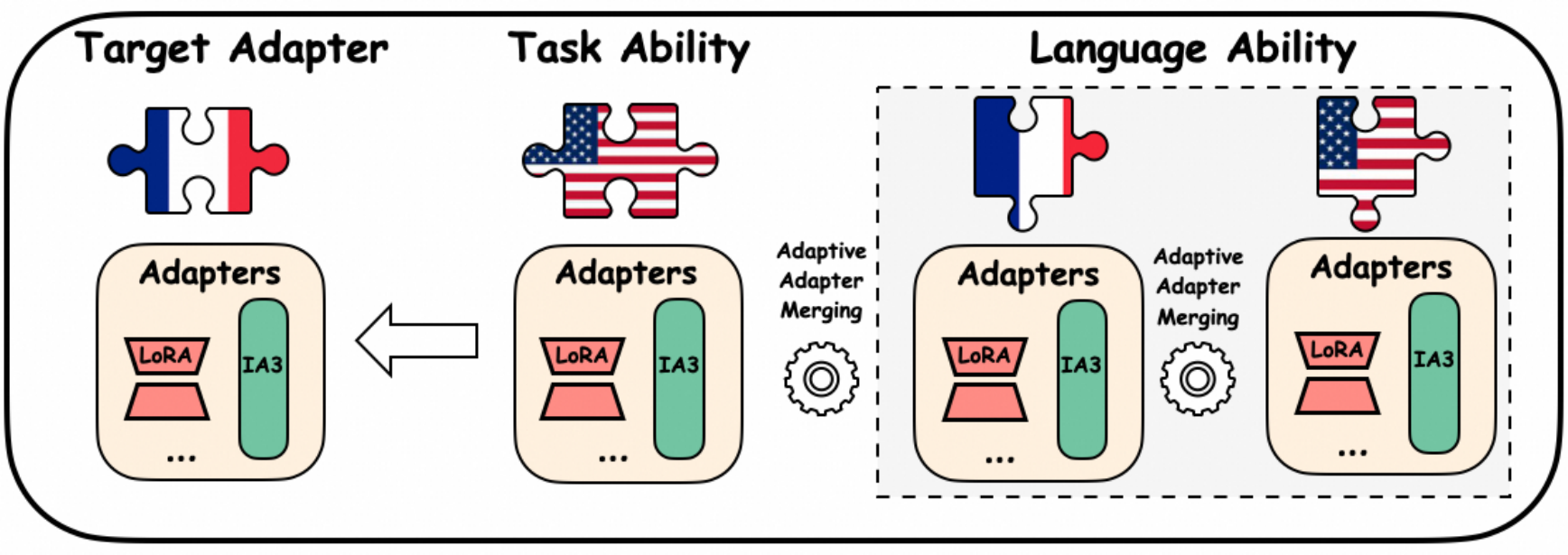}
  \caption{An overview of invariants of the language ability gap among different tasks in the adapter space, where by employing any three we can get the remaining one. In light of this observation, we propose \texttt{AdaMergeX}. } 
\label{fig:workflow}
\end{figure*}

On the contrary, some studies directly focus on enhancing the language abilities in target languages, so they endeavor to decouple task ability and language ability, enhance them separately, and subsequently merge them~\citep{pfeiffer2020mad, ansell2022composable, ponti2023combining}.
However, such an approach overlooks the intrinsic interdependence between task ability and language ability. Given that any specific task would be expressed in a particular language, these two abilities cannot be distinctly isolated from one another.

In this work, we argue that language ability and task ability are inherently interconnected. Instead of separating one from another, they should follow that task ability is affiliated with the source language while language ability refers to the capacity gap between the target language and the source language. In line with the famous equation ``$king - queen = man - woman$'' in the word embedding space~\citep{mikolov2013efficient}, we assume that the divergences between LLMs fine-tuned in different languages on a particular task follow the same distribution across diverse tasks. In the case of parameter-efficient fine-tuning, the equation becomes 
${read}^{\,fr}-{read}^{\,en}={math}^{\,fr}-{math}^{\,en}$
in the adapter space, where $read$ and $math$ refers to two tasks, and $fr$ and $en$ indicates two languages of the corresponding tasks. As shown in the left side of Figure \ref{fig:workflow}, in the adapter space, the divergence between the target language and source language on the target task follows the same distribution as the divergence on the reference task.

Therefore, we propose to accomplish the cross-lingual transfer through Adaptive Adapter Merging (\texttt{AdaMergeX}) with such a relation as shown in the right side of Figure \ref{fig:workflow}.
Specifically, we introduce a reference task from which we obtain the divergence between the target language and source language. Such a reference task can be an easily accessible task for both high-resource and low-resource languages, such as causal language modeling. In addition, we fine-tune LLMs on the target task in the source language. Finally, by merging the language ability and task ability, we can obtain the adapters of the target task in the target language.

Furthermore, in contrast to previous studies that combine models or adapters through a linear combination~\citep{ilharco2022editing, zhang2023composing, ponti2023combining}, we argue that the model merging method should align with the manner in which adapters are integrated with language models. Therefore, we design a structure-adaptive adapter merging method, which can adaptively select merging methods for LoRA~\citep{hu2021lora}, (IA)$^3$~\citep{liu2022few}, Adapter~\citep{houlsby2019parameter}, Prefix-Tuning~\citep{liliang2021prefix} etc. 

We evaluate the proposed \texttt{AdaMergeX} method on a wide range of multilingual tasks spanning 12 languages, covering a broad resource spectrum from high-resource to low-resource languages. Our evaluation demonstrates that \texttt{AdaMergeX} consistently outperforms other state-of-the-art methods including model merging, prompting, and general adapter merging methods. Notably, compared to MAD-X~\citep{pfeiffer2020mad} which separates the task and language ability with two adapters, \texttt{AdaMergeX} achieves $8.0\%$ and $15.9\%$ absolute improvement on XCOPA and XQuAD respectively with XLM-R. In the case of state-of-the-art adapter merging method Arimerge~\citep{zhang2023composing}, \texttt{AdaMergeX} achieves $31.1\%$ relative improvement on average in all languages and all tasks with Llama2. Moreover, the ablation analysis shows that \texttt{AdaMergeX} performs consistently well with different backbone models, source languages, and reference tasks.

\section{Background}\label{sec:back}

Given a pre-trained model, fine-tuning is often employed to improve the performance on specific tasks. Specifically, for a layer ${h} = {W}_0{x}$, where ${x}\in\mathbb{R}^k$ is input, ${h}\in\mathbb{R}^{d}$ is output and ${W}_0\in\mathbb{R}^{d\times k}$ is pre-trained parameters, fine-tuning updates parameters from ${W}_0$ to ${W}'$ and the layer becomes ${h} = {W}'{x}$. However, full fine-tuning requires many training data points and computing resources, which inspires the design of adapters~\citep{houlsby2019parameter}. With adapters, the layer is changed to ${h} = ({W}_0\circ {W}_A){x}$, where ${W}_A$ denotes the parameters of adapters and $\circ$ denotes the combination operation of pre-trained parameters and adapter parameters. 
During such parameter-efficient fine-tuning, pre-trained parameters ${W}_0$ are fixed and only adapter parameters ${W}_A$ are updated. 
With the number of parameters growing much bigger for LLMs, adapters become more widely used in the current practice of fine-tuning LLMs~\citep{hu2021lora, liliang2021prefix, liu2022few}

Various combination methods $\circ$ have been designed for different adapters.  
In this paper, we focus on two main widely used combination methods: addition and multiplication, corresponding to LoRA~\citep{hu2021lora} and (IA)$^3$~\citep{liu2022few}, respectively. We also involve Adapter~\citep{houlsby2019parameter} and Prefix-Tuning~\citep{liliang2021prefix} in to guarantee the generaliability.

\paragraph{LoRA} Specializing the combination method ``$\circ$'' to  element-wise addition denoted as ``$\oplus$'', LoRA employs low-rank decomposition to reduce training complexity. The layer is thus changed to 
\begin{equation}\label{equ:lora}
{h} = ({W}_0 \oplus {W}_A){x} = ({W}_0 \oplus {BA}){x},
\end{equation}
where ${B}\in\mathbb{R}^{d\times r}$ and ${A}\in\mathbb{R}^{r\times k}$ are low-rank decomposed matrices, and the rank $r\ll \min(d, k)$. Specifically, the LoRA can be implemented in any layer of the Transformer~\citep{vaswani2017attention} architecture, including the attention layer and the feed-forward layer.

\paragraph{(IA)$^3$} (IA)$^3$ specializes the combination method to element-wise multiplication ``$\odot$'': 
\begin{equation}\label{equ:ia3}
{h} = ({W}_0 \odot {W}_A) x,
\end{equation}
where $W_A\in \mathbb{R}^k$ is element-wise multiplied to each row of $W_0$. Furthermore, (IA)$^3$ can only be implemented to the key and value neuron in the attention layer and dimension reduction neuron in the feed-forward layer of the Transformer architecture.

\paragraph{Adapter \& Prefix-Tuning} By inserting layers and prefix tokens into the model, combination methods of Adapter and Prefix-Tuning can be formulated as
\begin{equation}\label{equ:comb}
{h} = ({W}_0 \| {W}_A) x,
\end{equation}
where $\|$ represents concatenation to original pre-trained parameters.

\section{AdaMergeX: Adaptive Adapter Merging for Cross-lingual Transfer}

\subsection{Cross-Lingual Transfer via Adapter Merging}\label{sec:cross-ling}

Generally, the ability of a model in a particular task and language can be seen as a composite of two abilities, namely, ``task ability'' and ``language ability''. The former denotes the model's competence in performing a certain task (e.g., text classification, sentence completion), whereas the latter signifies their general proficiency in the given language (e.g., English, Chinese, German).
Built on the premise that language ability and task proficiency are inherently intertwined, it is advocated that rather than isolating one from the other, the inference should be drawn that task ability is associated with the source language, whereas language ability refers to the capacity difference between the target language and the source language. In line with the famous equation
``$king - queen = man - woman$'' in the word embedding space, we assume
that the divergences between LLMs fine-tuned in different languages on a particular task follow the same distribution across diverse tasks.

Formally speaking, $A_{l_it_j}$ denotes the adapter of task $t_j$ in language $l_i$, then for any two languages $l_1$, $l_2$ and two NLP tasks $t_1$, $t_2$, we have
\begin{equation}\label{equ:samedis}
A_{l_{1}t_{1}} \| A_{l_{2}t_{1}} \sim A_{l_{1}t_{2}} \| A_{l_{2}t_{2}},
\end{equation}
where $ \| $ denotes the divergence among two adapters. 
For example, let's consider $l_1$ and $l_2$ as English and German, respectively, and $t_1$ and $t_2$ as the text classification task and question answering task, respectively. Assuming we have training data for each task in both languages, we can fine-tune LLMs to obtain four adapters: text classification in English, text classification in German, question answering in English, and question answering in German. We assume that the divergence between adapters for the text classification task in English and German, as well as the divergence between adapters for the question answering task in English and German, follows the same distribution. This divergence represents the  ``language ability'' that is independent of specific tasks. 

In the context of cross-lingual transfer, we aim to solve the task $t_1$ for the target language $l_1$, with the knowledge transferred from a source language $l_2$, which is often a high-resource language such as English. By imposing the condition of cross-lingual transfer, where labeled data is available only for the target task in the source language and there is unlabeled data in both the source and target languages, we can introduce another ``reference task'' $t_2$. This task can be easily constructed using unlabeled data, and language ability can be obtained by
$A_{l_{1}t_{2}} \| A_{l_{2}t_{2}}$. Moreover,
to obtain the ability of performing target task $t_1$ in the target language $l_1$, we can further transform Equation (\ref{equ:samedis}) as:
\begin{equation}\label{equ:spe}
A_{l_{1}t_{1}}  = A_{l_{2}t_{1}} \;\|^R\; (A_{l_{1}t_{2}} \| A_{l_{2}t_{2}}),
\end{equation}
where $\|^R$ is the reverse function of $\|$.
Intuitively, $A_{l_{2}t_{1}}$ represents the ``task ability'' in the source language, while $A_{l_{1}t_{2}} \| A_{l_{2}t_{2}}$ represents the ``language ability''. Through merging these two terms, we can transfer the ``task ability'' of $t_1$ from $l_2$ to $l_1$. 

To transfer the knowledge from labeled data in the high-resource language (i.e., given $A_{l_{2}t_{1}}$), the next step is to specify the reference task $t_2$. 
We observe that there are many easily obtained corpora of low-resource languages, such as Wikipedia, online blogs, etc. These corpora can be used to construct intuitive tasks such as causal language modeling, which can serve as the reference task $t_2$. Simultaneously, we can also construct such tasks for the high-resource language $l_2$. Therefore, adapters can be fine-tuned on such easily accessible reference tasks in different languages to obtain $A_{l_1t_2}$ and $A_{l_2t_2}$. Cross-lingual transfer thus can be achieved by merging these three adapters.

\subsection{Structure-Adaptive Adapter Merging}

As introduced in Section \ref{sec:back}, adapters have different structures, which inspires us to devise different adapter merging methods. We propose that the adapter merging approach must align with the way that the adapter combined with the original model. 

\paragraph{LoRA} In the fine-tuning process of LoRA, where the method involves element-wise addition to the original parameters, the merging method used to combine task ability and language ability should also employ element-wise addition. Additionally, since the divergence calculation approach $\|$ is intended to be the inverse function of the merging method, it should be carried out through element-wise subtraction in this scenario.
Therefore, Equation (\ref{equ:samedis}) is equivalently transferred to
\begin{equation}
    A_{l_{1}t_{1}} \ominus A_{l_{2}t_{1}} \sim A_{l_{1}t_{2}} \ominus A_{l_{2}t_{2}},
\end{equation}
where $\ominus$ denotes element-wise subtraction, and Equation (\ref{equ:spe}) is equivalently transferred to 
\begin{equation}\label{equ:merge_lora}
A_{l_{1}t_{1}}  = A_{l_{2}t_{1}} \oplus t\cdot(A_{l_{1}t_{2}} \ominus A_{l_{2}t_{2}}),
\end{equation}
where $\oplus$ denotes element-wise addition and $t$ is the hyper-parameter that adapts the scale of two distributions in the same family of distributions.

\paragraph{(IA)$^3$} Similarly, the fine-tuning method of (IA)$^3$ is element-wise multiplication to the original parameters, and the merging method should also be element-wise multiplication. Furthermore, we need to employ element-wise division to obtain the divergence between $A_{l_1t_2}$ and $A_{l_2t_2}$. Therefore, Equation (\ref{equ:samedis}) is equivalently transferred to
\begin{equation}
A_{l_{1}t_{1}} \oslash A_{l_{2}t_{1}} \sim A_{l_{1}t_{2}} \oslash A_{l_{2}t_{2}},
\end{equation}
where $\oslash$ denotes element-wise devision, and Equation (\ref{equ:spe}) is equivalently transferred to 
\begin{equation}\label{equ:merge_ia3}
A_{l_{1}t_{1}}  = A_{l_{2}t_{1}} \odot \Big( \big( t\cdot(A_{l_{1}t_{2}} \oslash A_{l_{2}t_{2}}) - \mathbbm{1}\big) + \mathbbm{1}\Big),
\end{equation}
where $\odot$ denotes element-wise multiplication and $t$ is the hyper-parameter determining the scale of two distributions in the same family of distributions.

\paragraph{Prefix-Tuning}
In the case of other adapter structures such as Prefix-Tuning, which involves the insertion of prefix tokens into the model, the merging process necessitates transferring adapters within the same space, such as MLP. Formally, the adaptive merging method is 
\begin{equation}\label{equ:merge_pre}
A_{l_1t_1} = t \cdot (A_{l_1t_2} * A_{l_2t_2}^{-1}) * A_{l_2t_1},    
\end{equation} where $*$ represents matrix multiplication and $A_{l_2t_2}^{-1}$ represents Moore-Penrose pseudo-inverse of the matrix. For Prefix-Tuning, $A_{lt}$ represents the prefix tokens. In this paper, we mainly focus on LoRA and (IA)$^3$ when Llama2 is the backbone model due to the subpar performance of prefix-tuning on fine-tuning~\citep{he2021towards}. On the contrary, in the case of smaller language models such as mT5~\citep{xue2021mt5}, we implement \texttt{AdaMergeX} on it with prefix-tuning. 
The experiment results are shown in Appendix \ref{apen:xlmr}.

\begin{table*}[t]
  \centering
\footnotesize
  \scalebox{1}{
  \begin{tabular}{lp{13cm}}
    \toprule
   \textbf{Task}  & \textbf{Zero-Shot Prompt} \\\midrule
   MGSM & Let's think step by step. Question: \{question\}  \\\midrule
   XCOPA  & Here is a premise and a question. Help me pick the more plausible option. Premise: \{premise\} Question: What is the \{question\}? (A) \{choice1\} (B) \{choice2\} \\\midrule
   XNLI  & You should judge whether the hypothesis is true (entailment), false (contradiction), or undetermined (neutral) given the premise. Premise: \{premise\} Hypothesis: \{hypothesis\} \\\midrule
   XQuAD  & \{context\} Question: \{question\} \\ \midrule
   XLSum  & Summarize the context in one sentence. Title: \{title\} Context: \{article\} \\
    \bottomrule
  \end{tabular}}
\caption{Zero-shot prompts for each dataset.}  \label{table:testset}
\end{table*}

\subsection{\texttt{AdaMergeX}}

Following notations in Section \ref{sec:cross-ling}, to solve a target task $t_1$ in a target language $l_1$, i.e., obtain the adapter $A_{l_{1}t_{1}}$, we need to fine-tune another three adapters: adapters on the target task in the source language ($A_{l_2t_1}$), adapters on the reference task in the target language ($A_{l_1t_2}$), and adapters on the reference task in the source language ($A_{l_2t_2}$). Note that $A_{l_1t_2}$ and $A_{l_2t_2}$ are easily obtainable, as we can choose any task in the target and source language. As mentioned earlier, the task can even be causal language modeling, which only requires unlabeled text corpora. Therefore, with only unlabeled data in both source and target language, our proposed \texttt{AdaMergeX} effectively transfers the target task proficiency from the source language to the target language. Moreover, given that the reference task remains constant, fine-tuning LLMs in the source language on the target task is the sole requirement for each new target task. This efficiency characterizes \texttt{AdaMergeX}.

In the case of LoRA, which fine-tunes LLMs by tuning $\{B, A\}$ in tuned layers of LLMs as introduced in Equation (\ref{equ:lora}), adapters are merged following Equation  (\ref{equ:merge_lora}) by element-wise addition and subtraction on $\{B, A\}$ in the corresponding layers of $A_{l_2t_1}$, $A_{l_1t_2}$, and $A_{l_2t_2}$. On the other hand, in the case of (IA)$^3$, the fine-tuning parameters are $W_A$ in tuned layers as depicted in Equation (\ref{equ:ia3}). Thus the merging method follows Equation (\ref{equ:merge_ia3}), which involves performing element-wise multiplication and division of the corresponding layers of $A_{l_2t_1}$, $A_{l_1t_2}$, and $A_{l_2t_2}$.

\section{Experiments}
\subsection{Experimental Setup}\label{sec:setup}

\paragraph{Datasets and Language}

To evaluate the effectiveness of our method, we conduct experiments on a wide variety of multilingual tasks in three main categories: reasoning tasks, natural language understanding (NLU) tasks, and natural language generation (NLG) tasks. For reasoning tasks, we test on multilingual arithmetic reasoning dataset XGSM~\citep{shi2022language} and multilingual commonsense reasoning dataset XCOPA~\citep{ponti2020xcopa}. For NLU tasks, we test on the multilingual natural language inference dataset XNLI~\citep{conneau2018xnli}, and question-answering dataset XQuAD~\citep{artetxe2020cross}. For NLG tasks, we test on multilingual summarization dataset XLSum~\citep{hasan2021xl}.
We choose $12$ languages that appear in more than once in the above datasets, including German (de), Russian (ru), French (fr), Spanish (es), Chinese (zh), Vietnamese (vi), Turkish (tr), Arabic (ar), Greek (el), Thai (th),  Hindi (hi), and Swahili (sw). 
Detailed settings of zero-shot prompts are shown in Table \ref{table:testset}. We utilize intuitive prompting methods for all tasks except for XCOPA and XNLI, where we employ prompts from~\citet{huang2023not}. Detailed examples of the prompting approach can be found in Appendix \ref{apen:prompt}. For MGSM, XCOPA and XQuAD, we adopt the whole testset, while for XNLI and XLSum we randomly sample $1000$ and $500$ data points from the whole testset respectively.

\paragraph{Baselines}

We conduct comparisons between our proposed method, which utilizes model merging for achieving cross-lingual transfer, and seven competing techniques: (i) Vanilla zero-shot prompting (``Vanilla''), which directly assesses target languages using the pre-trained LLM. (ii) English Tuning (``Eng-FT''), which involves fine-tuning the model in English for target tasks and subsequently transferring it directly to target languages. (iii) Cross-Lingual-Thought Prompting (``XLT (Vanilla)'')~\citep{huang2023not} achieves state-of-the-art results on cross-lingual transfer with LLMs through carefully designed prompt template, which involves explicit translation from the target to the source language, reasoning in the source language, and translating back to the target language. (iv) ``XLT (Eng-FT)'', where XLT approach is applied to the Eng-FT model.  (v) Arithmetic Merging (``AriMerge'')~\citep{zhang2023composing}, which is the state-of-the-art adapter merging method by arithmetic addition. (vi) MAD-X~\citep{pfeiffer2020mad} decomposes language and task via independent invertible adapters.  (vii) LF-SFT~\citep{ansell2022composable} adopts sparse fine-tuning on language and task respectively and directly merging via addition.

\begin{table*}[t]
  \centering
\footnotesize
  \scalebox{1}{
  \begin{tabular}{ll|cc|cc|c|l}
    \toprule
    \multirow{2}*{\textbf{{Adapters}}} & \multirow{2}*{\textbf{{Method}}} & \multicolumn{2}{c}{\textbf{\normalsize{Reasoning}}} \vline & \multicolumn{2}{c}{\textbf{\normalsize{NLU}}} \vline & \multicolumn{1}{c}{\textbf{\normalsize{NLG}}}\vline &  \multirow{2}*{\textbf{{Avg.}}}\\
 &  & MGSM & XCOPA & \multicolumn{1}{c}{XNLI}  & \multicolumn{1}{c}{XQuAD} \vline & \multicolumn{1}{c}{XLSum}\vline & \\
    \midrule
  \multirow{6}*{\begin{tabular}[c]{@{}l@{}} {{LoRA}} 
\end{tabular}}  & Vanilla & 2.7 & 52.3 & 14.8 & 0.0 & 20.9  & 18.1 \\
    & Eng-FT & 17.4 & 58.1 & 30.3 & 31.0 &  22.9  & 31.9 \\
 & XLT(Vanilla)& 2.8 & 52.6 & 23.7 & 19.3 & 1.3     &19.9 \\
  & XLT(Eng-FT)& 18.1 & 58.2 & 27.7 & 26.4 & 19.1 &29.9 \\
 & AriMerge & 6.0 & 57.9 & 13.6 & 30.1 & 19.5  & 25.4 \\
 & {\texttt{AdaMergeX}}  & \textbf{19.2} & \textbf{59.0} & \textbf{33.6}  & \textbf{31.6} & \textbf{23.3}  & \textbf{33.3}\\\midrule
  \multirow{6}*{\begin{tabular}[c]{@{}l@{}} {{(IA)$^3$}} 
\end{tabular}}  & Vanilla & 2.7 & 52.3 & 14.8 & 0.0 & 20.9  &18.1 \\
    & Eng-FT &  2.3 & 52.5 & 26.5  & 34.0 & 17.4  &26.5\\
 & XLT(Vanilla)& 2.8 & 52.6 & 23.7& 19.3  & 1.3  & 19.9 \\
  & XLT(Eng-FT)& 2.8 & 52.6 & 25.5 & 21.3 & 1.4  & 20.7 \\
 & AriMerge & 0.7 & 51.5 & 28.2 & 32.4  & 15.5 & 25.7 \\
 & {\texttt{AdaMergeX}}  & \textbf{3.9} & \textbf{53.1} & \textbf{28.6}  & \textbf{35.5} & \textbf{21.4} & \textbf{28.5}  \\
    \bottomrule  \end{tabular}}
\caption{Main experimental results on 5 representative cross-lingual tasks. Details of the selected zero-shot prompt, the baselines, and hyperparameters are described in Section \ref{sec:setup}.
}\label{table:result_main}
\end{table*}

\paragraph{Evaluation Metrics} For reasoning and NLU tasks, we use accuracy scores as our evaluation metric. For the summarization task, we evaluate the performance by ROUGE-L score \citep{lin2004rouge}. 

\paragraph{Experiment Details} The backbone model that we use to test \texttt{AdaMergeX} is Llama2-7b~\citep{touvron2023llama} for LoRA and (IA)$^3$, and XLM-R for Prefix-Tuning. To fine-tune Llama2 using LoRA and (IA)$^3$, we configure the target modules to include all available layers. We follow the notation of \cite{vaswani2017attention}. In particular, we utilize the attention layer's $\{W^Q, W^K, W^V, W^O\}$ and the feed-forward layer's $\{W_1, W_2\}$ for LoRA. For (IA)$^3$, we focus on $W^K$ and $W^V$ in the attention layer, as well as $W_2$ in the feed-forward layer. For the merging target modules, inspired by~\citet{gevaetal2021transformer} who attributes task ability to the feedword layer, we merge $\{W^Q, W^V\}$ for LoRA as we focus on language ability instead. We employ conventional causal language modeling as the reference task, where the prediction of the subsequent token is based on preceding inputs. Specifically, we generate the training set from the corpora provided by Wikimedia Foundation\footnote{\href{https://dumps.wikimedia.org/}{https://dumps.wikimedia.org/}} 
by dividing them into segments with a length of $512$. 
There is only one hyperparameter in our method, which is $t$ in Equation (\ref{equ:merge_lora}), (\ref{equ:merge_ia3}), and (\ref{equ:merge_pre}). When tuning this hyperparameter, for each task, we select the validation set from French and then extend it to encompass all other languages, for those tasks that do not contain French validation set, we adopt Vietnamese instead. 
For XLT method~\citep{huang2023not}, we adopt the same zero-shot prompts as in the original paper.

\subsection{Main Results}
Table \ref{table:result_main} presents our main experimental results on $5$ representative cross-lingual tasks with LlaMa2, where we report the average scores across all languages. Detailed results of each language are shown in Table \ref{table:result_lora} and \ref{table:result_ia3} in Appendix \ref{apen:dr} for LoRA and (IA)$^3$ respectively. Table \ref{table:xlmr} presents the results on XLM-R, where we compare with MAD-X and LF-SFT on XCOPA and XQuAD\footnote{We only test XCOPA and XQuAD because encoder-only models can only be applied to classification tasks.}. 

\begin{table*}[ht]
  \centering
  \footnotesize
  \scalebox{1.0}{
  \begin{tabular}{l|l|c|c|c|c|c|c|c}
    \toprule
    \textbf{Task} & \textbf{Method} & \textbf{tr} & \textbf{vi} & \textbf{th} & \textbf{sw}  & \textbf{el} & \textbf{ru} & \textbf{Avg.} \\
    \midrule
    \multirow{2}{*}{XCOPA} & MAD-X & 60.3 & 66.1 & 61.8 & 56.3  & - & -  & 59.5\\
    & \texttt{AdaMergeX} & 69.4 & 70.5 & 66.9 & 63.2 & - & -  & \textbf{67.5} \\
    \midrule
    \multirow{3}{*}{XQuAD} & MAD-X & - & - & 54.3 & 57.8 & 55.7 & 51.1  & 54.7 \\
    & LF-SFT & - & - & 65.5 & 64.6 & 75.2 & 58.6 & 66.0  \\
    & \texttt{AdaMergeX} & - & - & 70.2 & 70.4 & 77.9 & 63.8 & \textbf{70.6}  \\
    \bottomrule
  \end{tabular}}
  \caption{Experiment results on XCOPA and XQuAD with XLM-R, where AdaMergeX is implemented on LoRA.}\label{table:xlmr}
\end{table*}

\paragraph{\texttt{AdaMergeX} outperforms direct transfer and prompting methods} When comparing to fine-tuning on the task in English and direct transfer to the target language, \texttt{AdaMergX} outperforms it on all settings and achieves $1.4\%$ absolute improvement with LoRA and $1.5\%$ absolute improvement with (IA)$^3$. When comparing to the state-of-the-art method for cross-lingual transfer in LLMs via prompting, XLT with Vanilla Llama2 model (``XLT (Vanilla)'') and model fine-tuned on target task in English (``XLT (Eng-FT)''), \texttt{AdaMergeX} outperforms it on all settings and achieves $3.4\%$ absolute improvement with LoRA and $7.3\%$ absolute improvement with (IA)$^3$. This achievement proves that the introduction of adapter merging to achieve cross-lingual transfer is effective, especially in the circumstance of LLMs. 

\paragraph{\texttt{AdaMergeX} outperforms decoupling task ability and language ability method} As shown in Table \ref{table:xlmr}, compared to MAD-X and LF-SFT, which struggle to fully separate task ability from language ability, AdaMergeX demonstrates remarkable enhancements. In particular, AdaMergeX showcases an impressive absolute improvement of $8.0\%$ and $15.9\%$ on XCOPA and XQuAD, respectively, in comparison to MAD-X. Additionally, it achieves a significant $4.6\%$ absolute improvement on XQuAD when compared to LF-SFT. Therefore, our proposed new decoupling method is much more effective than others.

\paragraph{\texttt{AdaMergeX} outperforms general adapter merging methods} Compared with the state-of-the-art method for adapter merging namely Arimerge, \texttt{AdaMergeX} outperforms it on all settings and achieves $6.9\%$ absolute improvement with LoRA and $2.3\%$ absolute improvement with (IA)$^3$. Therefore, \texttt{AdaMergeX}, which adaptively considers the structure of adapters, outperforms all previous general adapter merging methods that adopt arithmetic addition for all kinds of adapters.

\paragraph{\texttt{AdaMergeX} performs consistently well with LoRA and (IA)$^3$} 

LoRA achieves higher absolute performance than (IA)$^3$, which shows the effectiveness of LoRA on fine-tuning. However, compared to the absolute improvement of \texttt{AdaMergeX} on LoRA and (IA)$^3$, they are comparable. For example, for MGSM, LoRA and (IA)$^3$ get the same absolute improvement $1.1\%$, and for XNLI, on which LoRA and (IA)$^3$ both achieve the highest absolute improvement, their performance are comparable. This proves that \texttt{AdaMergeX} performs consistently well on different adapters.

\subsection{Detailed Analysis}\label{sec:abla}

In this section, we validate the generalizability  of our proposed method across various aspects including the source language, reference task, backbone model, and target modules. Furthermore, we perform an ablation analysis to assess the essentiality of the adaptive merging method.

\paragraph{Source Language} To prove the generalizability of \texttt{AdaMergeX} on the source language, we explore its performance with different source languages in Table \ref{table:ablasl}. We test on five source languages including German, French, Spanish, Thai, and Vietnamese. We find that the performance is highly related to the source language, which depends on the language ability of the corresponding language. However, the improvements are consistent across languages. For example, the improvement was most significant with Vietnamese as the source language, with an absolute improvement of $3.4\%$ with LoRA and $3.8\%$ with (IA)$^3$. Therefore, \texttt{AdaMergeX} consistently performs well with different source languages.

\begin{table}[ht]
  \centering
\footnotesize
  \setlength{\tabcolsep}{1.7pt}
  \scalebox{0.9}{
  \begin{tabular}{ll|cc|cc|c|c}
    \toprule
   & \multirow{2}*{\textbf{{Method}}} & \multicolumn{2}{c}{\textbf{\normalsize{Reasoning}}} \vline & \multicolumn{2}{c}{\textbf{\normalsize{NLU}}} \vline & \multicolumn{1}{c}{\textbf{\normalsize{NLG}}}\vline &  \multirow{2}*{\textbf{{Avg.}}}\\
 &  & MGSM & XCOPA & \multicolumn{1}{c}{XNLI}  & \multicolumn{1}{c}{XQuAD} \vline & \multicolumn{1}{c}{XLSum}\vline & \\
    \midrule
\multirow{12}*{\rotatebox{90}{LoRA}}&  De-Tune & 20.9 & $-$ & 48.3 & 44.4 & $-$ & 37.9 \\
  &  {\texttt{AdaMergeX}}  & 22.3 & $-$ & 50.9 & 46.5 & $-$ & \textbf{39.9} \\    \cmidrule{2-8}
&  Fr-Tune & 19.9 & $-$ & 52.9 & $-$ & 24.1 & 32.3  \\
  &  {\texttt{AdaMergeX}}  & 22.2 & $-$ & 57.1 & $-$ & 24.8 & \textbf{34.7}  \\\cmidrule{2-8}
&  Es-Tune & 19.2 & $-$ & 33.9 & 45.4 & 22.1 & 30.2 \\
  &  {\texttt{AdaMergeX}}  & 18.7 & $-$ & 35.1 & 49.1 & 23.7 & \textbf{31.7} \\\cmidrule{2-8}
& Th-Tune & 3.2 & 49.3 & 1.9 & 39.8 & 20.3 & 22.9 \\
  &  {\texttt{AdaMergeX}}  & 4.5 & 48.9 & 6.2 & 44.2 & 20.1 & \textbf{24.8} \\\cmidrule{2-8}
&  Vi-Tune & $-$ & 63.8 & 49.1 & 36.2 & 21.7 & 42.7 \\
  &  {\texttt{AdaMergeX}}  & $-$ & 64.2 & 53.2 & 38.9 & 22.3 & \textbf{44.7}
  \\\midrule
\multirow{12}*{\rotatebox{90}{(IA)$^3$}}&  De-Tune & 2.9 & $-$ & 43.5 & 45.6 & $-$ & 30.7 \\
  &  {\texttt{AdaMergeX}}  & 6.3 & $-$ & 44.0 & 47.1 & $-$ & \textbf{32.5} \\   \cmidrule{2-8}
&  Fr-Tune & 2.5 & $-$ & 48.7 & $-$ & 19.8 & 23.7 \\
  &  {\texttt{AdaMergeX}}  & 4.1 & $-$ & 47.9 & $-$ & 21.6 & \textbf{24.5} \\\cmidrule{2-8}
& Es-Tune & 3.5 & $-$ & 49.2 & 45.9 & 18.2 & 29.2 \\
  &  {\texttt{AdaMergeX}}  & 5.3 & $-$ & 50.9 & 44.6 & 20.1 & \textbf{30.2} \\\cmidrule{2-8}
&  Th-Tune & 1.2 & 49.8 & 0.0 & 27.7 & 20.2 & 19.8 \\
  &  {\texttt{AdaMergeX}}  & 1.9 & 50.4 & 0.0 & 28.9 & 24.1 & \textbf{21.1} \\\cmidrule{2-8}
&  Vi-Tune & $-$ & 49.8 & 45.5 & 33.2 & 20.1 & 37.2 \\
  &  {\texttt{AdaMergeX}}  & $-$ & 48.7 & 50.2 & 36.1 & 22.5 & \textbf{39.4}
  \\
    \bottomrule
  \end{tabular}}
\caption{Ablation study on source language.}
\label{table:ablasl}
\end{table}

\paragraph{Reference Task}

To prove the generalizability of \texttt{AdaMergeX} on the reference task, we explore its performance with different reference task in Table \ref{table:ablast}. We test on three different reference tasks, including XCOPA, XNLI, XQuAD, while the source language is English. The dataset was tested on the corresponding available languages among German, French, Spanish, Thai, and Vietnamese.  Specifically, the improvement was most significant with XQuAD as the reference task, with an absolute improvement of $1.3\%$ with LoRA and $1.7\%$ with (IA)$^3$.  Thus, it verifies that \texttt{AdaMergeX} is general to any reference task.

\begin{table}[ht]
  \centering
\footnotesize
  \setlength{\tabcolsep}{1.5pt}
  \scalebox{0.8}{
  \begin{tabular}{lll|cccccc}
    \toprule
  & \multirow{1}*{\textbf{{Ref. Task}}} & \multirow{1}*{\textbf{{Method}}} & \textbf{MGSM} & \textbf{XCOPA} & \textbf{XNLI} & \textbf{XQuAD}  & \textbf{XLSum} & \textbf{Avg.}  \\
    \midrule
\multirow{6}*{\rotatebox{90}{LoRA}} & $-$ & Eng-Tune & 14.4  & 59.9 & 44.6 & 42.3 & 16.1 & 35.1 \\ \cmidrule{2-9}
  & XCOPA & {\texttt{AdaMergeX}}  & 15.2  & 60.2 & 45.1 & 43.8 & 18.2 & 36.5 
  \\    \cmidrule{2-9}
  & XNLI & {\texttt{AdaMergeX}}  & 14.5  & 60.9 & 46.7 & 44.1 & 18.4 & \textbf{36.9} 
  \\    \cmidrule{2-9}
  & XQuAD & {\texttt{AdaMergeX}}  & 14.9  & 61.8 & 45.4 & 44.4 & 18.1 & \textbf{36.9}  
 \\\midrule
\multirow{6}*{\rotatebox{90}{(IA)$^3$}} & $-$ & Eng-Tune & 2.6  & 52.7 & 40.0 & 39.2 & 10.8 & 29.1  \\ \cmidrule{2-9}
  & XCOPA & {\texttt{AdaMergeX}}  & 4.9  & 54.3 & 40.5 & 40.4 & 12.4 & 30.5   
  \\    \cmidrule{2-9}
  & XNLI & {\texttt{AdaMergeX}}  & 3.6 & 54.6 & 41.2 & 39.9 & 13.1 & 30.5  
  \\    \cmidrule{2-9}
  & XQuAD & {\texttt{AdaMergeX}}  & 4.1  & 53.9 & 42.1 & 41.0 & 12.9 & \textbf{30.8} 
  \\
    \bottomrule
  \end{tabular}}
\caption{Ablation study on reference Task. }
\label{table:ablast}
\end{table}

\paragraph{Backbone Models}

Not limited to Decode-only Models such as Llama2, we do further analysis on Encoder-Decoder model T5-base~\citep{raffel2020exploring} to prove its universal effectiveness. \texttt{AdaMergeX} achieves consistently the best performance compared to fine-tuning on English and AriMerge as shown in Table \ref{table:t5} of Appendix \ref{apen:t5}. Furthermore, we also implement our method on Encoder-only model XLM-R and compare with MAD-X and LF-SFT as shown in Table \ref{table:xlmr}. This shows the flexibility of choosing the backbone model when implementing \texttt{AdaMergeX}.

\paragraph{Merging Method}

We conduct an ablation analysis on merging method to ascertain the indispensability and the effectiveness of adaptive merging in \texttt{AdaMergeX}. Table \ref{table:abla_mer} in Appendix \ref{apen:adaptive} shows the detailed results, where \texttt{AdaMergeX} (adaptive) represents \texttt{AdaMergeX} with adaptive merging methods, while \texttt{AdaMergeX} (cross) represents \texttt{AdaMergeX} with cross merging methods, i.e., LoRA with merging method of (IA)$^3$ and vice versa. We find that when applying the merging method of (IA)$^3$ to LoRA, the performance is reduced much, and vice versa. As a result, the adaptive merging method is crucial for adapter merging.

\section{Related Work}

\subsection{Cross-Lingual Transfer}
The emergence of multilingual systems~\citep{kenton2019bert, conneau2019cross, conneau2020unsupervised, chatgpt, anil2023palm, touvron2023llama} has sparked interest in cross-lingual transfer~\citep{kim2017cross, lin2019choosing, schuster2019cross, pfeiffer2020mad}. Fine-tuning on the target language and target task is an intuitive way to make models obtain the ability of this task, but it is too costly in the era of LLMs as we always lack enough training data~\citep{ma2023prompt}. Alternatively, some researchers explore realigning representations among languages~\citep{nguyen2023enhancing, salesky2023pixel, gao2023improving}. However, \citet{gaschi2023exploring} demonstrates that aligned representations do not significantly benefit cross-lingual transfer. To address this issue, some works adopt explicit translation to achieve cross-lingual transfer~\citep{liang2023machine, huang2023not}. However, they rely on translation ability which is not guaranteed. In addition, \citet{pfeiffer2020mad} and \citet{ansell2022composable} decouple language ability and task ability, but they ignore the interconnection of these two abilities. Furthermore, in the era of in-context learning~\citep{brown2020language, chowdhery2022palm, touvron2023llama, openai2023gpt4}, \citet{li2023enhancing} and \citet{tanwar2023multilingual} utilize prompt tuning to achieve cross-lingual transfer. Nevertheless, the performance remains limited for low-resource languages, which is often not carefully considered in the pre-training of LLMs.

\subsection{Model Merging}
Model merging has been widely used in image identification~\citep{wortsman2022model, matena2022merging}, knowledge editing~\citep{mitchell2022memory, meng2022mass} and task combination~\citep{ilharco2022editing}. In the era of PEFT, researchers have started exploring different approaches to merging adapters~\citep{zhang2023composing, yadav2023resolving, huang2023lorahub, chronopoulou2023adaptersoup, ponti2023combining}. These studies, however, have primarily focused on task transfer and have solely utilized linear combinations of different adapters, which may not be applicable to all types of adapters. Moreover, the utilization of model merging for cross-lingual transfer is under-studied.

\section{Conclusion}

In this work, we propose a new cross-lingual transfer method \texttt{AdaMergeX}. We split target task ability in the target language into two parts: ``task ability'' and ``language ability''. In the context of PEFT, task ability can be obtained by tuning on the target task in the source language. To achieve cross-lingual transfer, which aims to transfer task ability from the source language to the target language, we introduce a reference task from which we obtain language ability and further merge it to task ability by adapter merging. Different from all previous adapter merging methods, we propose a structure adaptive adapter merging method that aligns the adapter merging method with the way adapters combined to LLMs. Experiment results show that \texttt{AdaMergeX} performs well among all settings. Moreover, ablation analysis proves that \texttt{AdaMergeX} is robust to backbone models, source languages, and source tasks.

\section*{Limitations}

Our research primarily utilizes models with around 7 billion parameters, specifically Llama2-7b, due to limitations in computational resources. Exploring our methodologies on larger-scale models may offer further valuable perspectives. Furthermore, although the training set for the reference task is easily accessible, fine-tuning the parameters of the entire model necessitates a certain investment of time. However, this training time can be significantly reduced by integrating language-specific adapters or employing language-specific Mixture of Experts (MoE) techniques, which ultimately lowers the overall training cost.


\bibliography{anthology,custom}
\bibliographystyle{acl_natbib}

\appendix
\section{Appendix}

\subsection{\texttt{AdaMergeX} on Prefic-Tuning}\label{apen:xlmr}

The results demonstrate that $\texttt{AdaMergeX}$ excels remarkably within the realm of prefix-tuning, a distinct and separate approach to fine-tuning. Results on XNLI task with mT5~\citep{xue2021mt5} are shown as follows in Table \ref{apen:prefix}.

\begin{table*}[ht]
  \centering
  \footnotesize
  \begin{tabular}{l|l|c|c|c|c|c|c|c|c|c}
    \toprule
    \textbf{Task} & \textbf{Method} & \textbf{es} & \textbf{fr} & \textbf{ru}  & \textbf{tr} & \textbf{vi} & \textbf{th} & \textbf{sw}  & \textbf{el}  & \textbf{Avg.} \\
    \midrule
    \multirow{2}{*}{XCOPA} & Eng-FT & $-$ & $-$  & $-$  & $-$ & 69.5 & 57.4 & 62.8 & $-$ & 65.2 \\
    & AriMerge  & $-$ & $-$  & $-$  & $-$ & 65.4 & 59.7 & 64.1 & $-$ & 63.1 \\
    & AdaMergeX & $-$ & $-$  & $-$  & $-$ & 71.3 & 63.2 & 65.6 & $-$ & \textbf{66.7} \\
    \midrule
    \multirow{3}{*}{XNLI} & Eng-FT & 31.2 & 29.7 & 30.4 & 19.8 & 43.1 & 11.6 & 13.2 & 16.3 & 24.4 \\
    & AriMerge  & 29.8 & 28.3 & 33.2 & 21.4 & 42.9 & 11.8  & 14.6 & 21.8 & 25.5  \\
    & AdaMergeX  & 34.1 & 31.4 & 34.2 & 20.9 & 44.8 & 20.3  & 16.7 & 25.3 & \textbf{28.5}  \\\midrule
    \multirow{3}{*}{XLSum} & Eng-FT & 13.4 & 14.2 & 12.7  & 14.1 & 18.9 & 14.9 & 7.8 & $-$ & 13.7   \\
    & AriMerge  & 14.5 & 15.2 & 15.6  & 13.9 & 20.2 & 15.6 & 8.6  & $-$ & 14.8 \\
    & AdaMergeX  & 14.9 & 16.1 & 17.4 & 16.1 & 19.8 & 17.1 & 10.3  & $-$ & \textbf{16.0}  \\
    \bottomrule
  \end{tabular}
  \caption{Results of \texttt{AdaMergeX} on Prefix-tuning with mT5.}\label{apen:prefix}

\end{table*}

\subsection{Prompts}\label{apen:prompt}

Detailed prompts of tasks in each language are listed in Figure \ref{fig:prompt_extract}.

\begin{figure*}[t]
\centering
\includegraphics[width=0.8\textwidth]{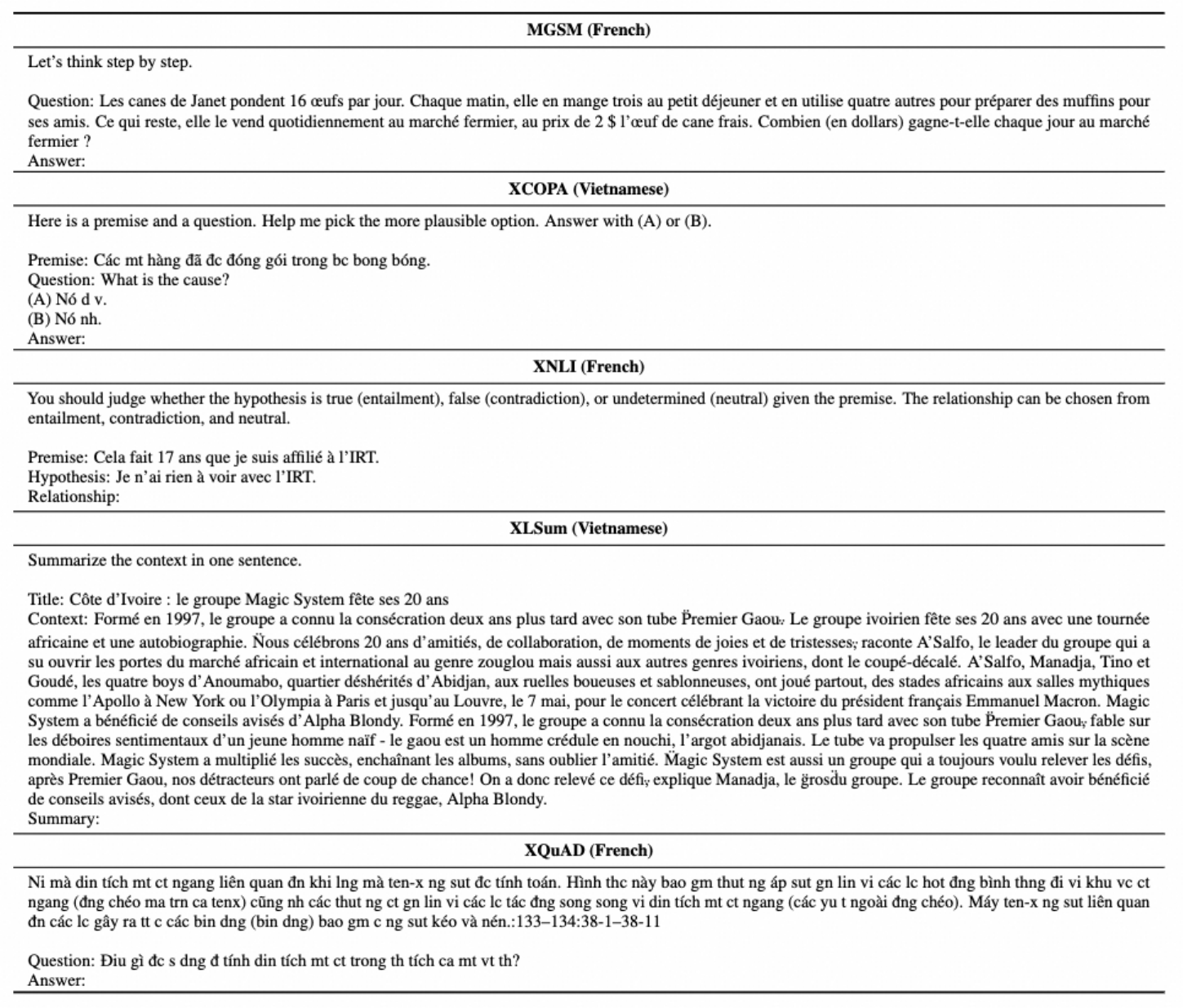}
  \caption{One-shot prompting examples of tested datasets. } 
\label{fig:workflow}
\end{figure*}

\subsection{Detailed Results}\label{apen:dr}

We present detailed results in Table \ref{table:result_lora} and Table \ref{table:result_ia3}.

\begin{table*}[ht]
  \centering
\caption{Comprehensive experimental results for both baselines and \texttt{AdaMergeX} are obtained across all datasets in corresponding available languages. The fine-tuning method employed was LoRA, with Llama2-7b serving as the backbone model.}
\footnotesize
  \scalebox{0.9}{
  \begin{tabular}{ll|llllllllllll}
    \toprule
  \multirow{1}*{\textbf{{Models}}} & \multirow{1}*{\textbf{{Method}}} & de & ru & fr & es & zh & vi & tr & ar & el & th & hi & sw \\
    \midrule
   \multirow{3}*{\begin{tabular}[c]{@{}l@{}} {{MGSM}} 
\end{tabular}} & Vanilla & 2.4 & 3.6 & 3.6 & 3.2 & 2.4 & $-$ & $-$ & $-$ & $-$ & 2.0 & $-$ & 2.0 \\
   & Eng-FT & 22.4 & 24.8 & 20.4 & 22.4 & 22.8 & $-$ & $-$ & $-$ & $-$ & 6.8 & $-$ & 2.4 \\
& XLT(Vanilla) & 2.0 & 2.8 & 2.8 & 3.2 & 2.8 & $-$ & $-$ & $-$ & $-$ & 2.0 & $-$ & 3.2 \\
& XLT(Eng-FT) & 22.0 & 24.0 & 22.8 & 24.4 & 24.2 & $-$ & $-$ & $-$ & $-$ & 5.2 & $-$ & 4.4 \\
  & AriMerge & 6.4 & 8.0 & 2.4 & 10.4 & 3.2 & $-$ & $-$ & $-$ & $-$ & 11.6 & $-$ & 0.0 \\
   & {\texttt{AdaMergeX}}  & 24.8 & 26.2 & 23.6 & 22.4 & 22.0 & $-$ & $-$ & $-$ & $-$ & 8.0 & $-$ & 7.2\\\midrule
 \multirow{3}*{\begin{tabular}[c]{@{}l@{}}{{XCOPA}}
\end{tabular}} & Vanilla & $-$ & $-$ & $-$ & $-$ & 54.4 & 54.0 & $-$ & $-$ & $-$ & 51.8 & $-$ & 49.0 \\
   & Eng-FT & $-$ & $-$ & $-$ & $-$ & 61.8 & 67.2 & $-$ & $-$ & $-$ & 52.6 & $-$ & 50.6 \\
& XLT(Vanilla) & $-$ & $-$ & $-$ & $-$ & 56.8 & 52.4 & $-$ & $-$ & $-$ & 51.0 & $-$ & 50.0 \\
& XLT(Eng-FT) & $-$ & $-$ & $-$ & $-$ & 60.6 & 70.0 & $-$ & $-$ & $-$ & 51.6 & $-$ & 50.4 \\
  & AriMerge & $-$ & $-$ & $-$ & $-$ & 61.0 & 69.8 & $-$ & $-$ & $-$ & 50.6 & $-$ & 50.0 \\
   & {\texttt{AdaMergeX}}  & $-$ & $-$ & $-$ & $-$ & 61.8 & 69.8 & $-$ & $-$ & $-$ & 51.8 & $-$ & 52.2\\\midrule
 \multirow{3}*{\begin{tabular}[c]{@{}l@{}}{{XNLI}}
\end{tabular}}  & Vanilla & 27.4 & 26.6 & 24.0 & 20.2 & 0.3 & 21.5 & 14.3 & 0.1 & 0.3 & 0.3 & 0.0 & 43.0 \\
   & Eng-FT & 54.0 & 54.0 & 58.2 & 60.5 & 33.5 & 47.0 & 9.6 & 0.8 & 5.4 & 3.3 & 5.2 & 31.8 \\
  & XLT(Vanilla) & 44.7 & 44.4 & 39 & 36.9 & 5.3 & 36 & 20.6 & 0.4 & 0.2 & 13.9 & 0.2 & 42.6 \\
& XLT(Eng-FT) & 54.1 & 44.3 & 44.6 & 58.6 & 34.0 & 43.0 & 15.9 & 0.0 & 1.2 & 2.0 & 0.9 & 33.9 \\
  & AriMerge & 28.7 & 16.5 & 12.8 & 21.2 & 1.0 & 32.1 & 16.2 & 0.3 & 1.8 & 0.0 & 10.2 & 22.8 \\
   & {\texttt{AdaMergeX}}  & 57.8 & 56.7 & 63.1 & 62.8 & 32.9 & 49.2 & 10.3 & 1.0 & 9.1 & 13.3 & 14.9 & 35.9\\\midrule
 \multirow{3}*{\begin{tabular}[c]{@{}l@{}}{{XLSum}}
\end{tabular}} & Vanilla & $-$ & 13.4 & 12.5 & 11.4 & 56.0 & 22.1 & 15.7 & 23.5 & $-$ & 14.8 & 31.6 & 8.1 \\
   & Eng-FT & $-$ & 21.7 & 16.1  & 11.3 & 58.4 & 21.2 & 16.4 & 25.8 & $-$ & 15.6 & 32.9 & 9.9 \\
& XLT(Vanilla) & $-$ & 0.6 & 2.3 & 1.8 & 0.5 & 1.3 & 2.5 & 0.8 & $-$ & 0.2 & 0.8 & 2.1 \\
& XLT(Eng-FT) & $-$ & 17.8 & 5.0 & 6.6 & 56.8 & 13.5 & 10.8 & 28.9 & $-$ & 13.5 & 33.9 & 3.9 \\
  & AriMerge & $-$ & 14.5 & 8.7 & 9.8 & 49.8 & 12.6 & 11.7 & 29.8 & $-$ & 17.2 & 34.2 & 6.5 \\
   & {\texttt{AdaMergeX}}  & $-$ & 21.6 & 16.2 & 11.9 & 58.4 & 21.6 & 16.7 & 25.6 & $-$ & 15.5 & 33.9 & 11.4 \\\midrule
 \multirow{3}*{\begin{tabular}[c]{@{}l@{}}{{XQuAD}}
\end{tabular}}  & Vanilla & 0.0 & 0.0 & $-$ & 0.0 & 0.0 & 0.0 & 0.0 & 0.0 & 0.0 & 0.0 & 0.0 & $-$ \\
   & Eng-FT & 49.0 & 34.1 & $-$ & 48.2 & 53.5 & 40.9 & 17.3 & 10.2 & 13.9 & 31.0 & 11.8 & $-$ \\
& XLT(Vanilla) & 34.8 & 14.0 & $-$ & 29.8 & 33.1 & 21.8 & 20.2 & 12.0 & 8.6 & 7.1 & 12.1 & $-$ \\
& XLT(Eng-FT) & 39.1 & 26.3 & $-$ & 40.7 & 41.2 & 33.9 & 19.0 & 13.8 & 13.0 & 23.8 & 13.2 & $-$ \\
  & AriMerge & 50.7 & 31.8 & $-$ & 49.1 & 50.2 & 42.3 & 15.9 & 10.4 & 12.6 & 28.7 & 9.7 & $-$ \\
   & {\texttt{AdaMergeX}}  & 50.7 & 34.1 & $-$ & 50.0 & 53.2 & 41.7 & 17.3 & 10.4 & 13.7 & 31.8 & 13.1 & $-$ \\
    \bottomrule
  \end{tabular}}
  \label{table:result_lora}
\end{table*}

\begin{table*}[ht]
  \centering
\caption{Comprehensive experimental results for both baselines and \texttt{AdaMergeX} are obtained across all datasets in corresponding available languages. The fine-tuning method employed was (IA)$^3$, with Llama2-7b serving as the backbone model.}
\footnotesize
  \scalebox{0.9}{
  \begin{tabular}{ll|llllllllllll}
    \toprule
  \multirow{1}*{\textbf{{Models}}} & \multirow{1}*{\textbf{{Method}}} & de & ru & fr & es & zh & vi & tr & ar & el & th & hi & sw \\
    \midrule
   \multirow{3}*{\begin{tabular}[c]{@{}l@{}} {{MGSM}} 
\end{tabular}} & Vanilla & 2.4 & 3.6 & 3.6 & 3.2 & 2.4 & $-$ & $-$ & $-$ & $-$ & 2.0 & $-$ & 2.0 \\
   & Eng-FT & 2.0 & 2.0 & 3.6 & 2.4 & 1.6 & $-$ & $-$ & $-$ & $-$ & 2.4 & $-$ & 2.0 \\
& XLT(Vanilla) & 2.0 & 2.8 & 2.8 & 3.2 & 2.8 & $-$ & $-$ & $-$ & $-$ & 2.0 & $-$ & 3.2 \\
& XLT(Eng-FT) & 0.8 & 1.6 & 4.8 & 4.0 & 3.2 & $-$ & $-$ & $-$ & $-$ & 2.8 & $-$ & 2.4 \\
& AriMerge & 0.0 & 0.4 & 0.4 & 0.0 & 1.6 & $-$ & $-$ & $-$ & $-$ & 2.0 & $-$ & 0.4 \\
   & {\texttt{AdaMergeX}}  & 4.4 & 3.6 & 4.8 & 6.0 & 3.6 & $-$ & $-$ & $-$ & $-$ & 2.8 & $-$ & 2.0 \\\midrule
 \multirow{3}*{\begin{tabular}[c]{@{}l@{}}{{XCOPA}}
\end{tabular}} & Vanilla & $-$ & $-$ & $-$ & $-$ & 54.4 & 54.0 & $-$ & $-$ & $-$ & 51.8 & $-$ & 49.0 \\
   & Eng-FT & $-$ & $-$ & $-$ & $-$ & 54.8 & 54.2 & $-$ & $-$ & $-$ & 51.2 & $-$ & 49.8 \\
& XLT(Vanilla) & $-$ & $-$ & $-$ & $-$ & 56.8 & 52.4 & $-$ & $-$ & $-$ & 51.0 & $-$ & 50.0 \\
& XLT(Eng-FT) & $-$ & $-$ & $-$ & $-$ & 56.8 & 53.2 & $-$ & $-$ & $-$ & 51.4 & $-$ & 49.8 \\
& AriMerge & $-$ & $-$ & $-$ & $-$ & 53.0 & 50.6 & $-$ & $-$ & $-$ & 52.2 & $-$ & 50.2\\
   & {\texttt{AdaMergeX}}  & $-$ & $-$ & $-$ & $-$ & 55.0 & 55.2 & $-$ & $-$ & $-$ & 52.1 & $-$ & 50.0
   \\\midrule
 \multirow{3}*{\begin{tabular}[c]{@{}l@{}}{{XNLI}}
\end{tabular}} & Vanilla & 27.4 & 26.6 & 24.0 & 20.2 & 0.3 & 21.5 & 14.3 & 0.1 & 0.3 & 0.3 & 0.0 & 43.0 \\
   & Eng-FT & 46.4 & 45.3 & 51.9 & 50.7 & 1.6 & 51.0 & 31.4 & 0.1 & 0.8 & 0.0 & 0.0 & 39.3 \\
& XLT(Vanilla) & 44.7 & 44.4 & 39.0 & 36.9 & 5.3 & 36.0 & 20.6 & 0.4 & 0.2 & 13.9 & 0.2 & 42.6 \\
& XLT(Eng-FT) & 34.3 & 36.8 & 36.3 & 34.2 & 25.4 & 34.4 & 32.1 & 5.2 & 3.8 & 20.7 & 8.0 & 34.4 \\
& AriMerge & 42.4 & 47.2 & 52.9 & 49.3 & 6.4 & 54.5 & 49.1 & 0.2 & 0.5 & 0.1 & 0.0 & 35.5 \\
   & {\texttt{AdaMergeX}}  & 45.3 & 46.5 & 53.0 & 54.3 & 1.5 & 58.8 & 41.7 & 2.2 & 0.9 & 0.1 & 0.1 & 38.4\\\midrule
 \multirow{3}*{\begin{tabular}[c]{@{}l@{}}{{XLSum}}
\end{tabular}} & Vanilla & $-$ & 13.4 & 12.5 & 11.4 & 56.0 & 22.1 & 15.7 & 23.5 & $-$ & 14.8 & 31.6 & 8.1 \\
   & Eng-FT & $-$ & 4.2 & 9.0 & 6.8 & 56.6 & 14.7 & 13.6 & 16.6 & $-$ & 12.5 & 32.3 & 7.6 \\
& XLT(Vanilla) & $-$ & 0.6 & 2.3 & 1.8 & 0.5 & 1.3 & 2.5 & 0.8 & $-$ & 0.2 & 0.8 & 2.1 \\
& XLT(Eng-FT) & $-$ & 0.6 & 3.1 & 1.8 & 0.4 & 1.3 & 2.5 & 1.1 & $-$ & 0.3 & 0.8 & 2.1 \\
& AriMerge & $-$ & 4.8 & 6.3 & 7.6 & 44.1 & 9.9 & 11.8 & 15.4 & $-$ & 13.1 & 32.3 & 9.4 \\
   & {\texttt{AdaMergeX}}  & $-$ & 14.5 & 13.1 & 11.5 & 55.2 & 24.4 & 15.3 & 23.5 & $-$ & 13.6 & 33.4 & 9.2 \\\midrule
 \multirow{3}*{\begin{tabular}[c]{@{}l@{}}{{XQuAD}}
\end{tabular}} & Vanilla & 0.0 & 0.0 & $-$ & 0.0 & 0.0 & 0.0 & 0.0 & 0.0 & 0.0 & 0.0 & 0.0 & $-$ \\
   & Eng-FT & 47.3 & 32.8 & $-$ & 47.6 & 53.7 & 35.1 & 28.9 & 22.8 & 21.9 & 26.9 & 23.2 & $-$ \\
& XLT(Vanilla) & 34.8 & 14.0 & $-$ & 29.8 & 33.1 & 21.8 & 20.2 & 12.0 & 8.6 & 7.1 & 12.1 & $-$ \\
& XLT(Eng-FT) & 37.1 & 16.8 & $-$ & 32.4 & 37.6 & 25.1 & 19.3 & 14.0 & 10.0 & 7.0 & 14.1 & $-$ \\
& AriMerge & 46.0 & 32.2 & $-$ & 44.5 & 51.2 & 35.4 & 28.2 & 23.4 & 20.6 & 21.6 & 20.7 & $-$ \\
   & {\texttt{AdaMergeX}}  & 48.6 & 33.0 & $-$ & 48.2 & 56.0 & 35.7 & 29.3 & 25.4 & 24.5 & 29.2 & 24.6 & $-$ \\
    \bottomrule
  \end{tabular}}
  \label{table:result_ia3}
\end{table*}

\subsection{\texttt{AdaMergeX} on T5-base}\label{apen:t5}

Because T5-base only supports Spanish and French in chosen languages, we only test these two languages. In the case of LoRA on XNLI, \texttt{AdaMergeX} obtains $4.2\%$ absolute improvements in Spanish and $2.8\%$ absolute improvements in French. For (IA)$^3$, the improvements are $1.1\%$ and $4.0\%$ respectively. 

\begin{table*}[hbpt]
  \caption{Ablation study on backbone models. Results are evaluated on T5-base.}
  \centering
    \footnotesize
  \begin{tabular}{ll|l|c|c|c}
    \toprule
  \textbf{Adapters} & \textbf{Task} & \textbf{Method} & \textbf{es} & \textbf{fr} & \textbf{Avg.} \\\midrule
  \multirow{6}*{LoRA}
    & \multirow{3}{*}{XNLI} & Eng-FT & $33.0$ & $32.9$ & $33.0$ \\
    & & AriMerge  & $34.1$ & $30.1$ & $32.1$  \\
    & & AdaMergeX  & $37.2$ & $35.7$ & $36.5$  \\\cmidrule{2-6}
    & \multirow{3}{*}{XLSum} & Eng-FT & $12.4$ & $15.3$ & $13.9$ \\
    & & AriMerge  & $13.1$ & $16.5$ & $14.8$  \\
    & & AdaMergeX  & $14.9$ & $16.6$ & $15.8$ \\
    \midrule
  \multirow{6}*{(IA)$^3$}
    & \multirow{3}{*}{XNLI} & Eng-FT & $38.2$ & $38.4$ & $38.3$  \\
    & & AriMerge  & $35.6$ & $36.1$ & $35.9$  \\
    & & AdaMergeX  & $39.3$ & $42.4$ & $40.8$   \\\cmidrule{2-6}
    & \multirow{3}{*}{XLSum} & Eng-FT & $13.2$ & $14.7$ & $14.0$  \\
    & & AriMerge  & $14.3$ & $15.1$ & $14.7$  \\
    & & AdaMergeX  & $14.2$ & $16.7$ & $15.5$  \\
    \bottomrule
  \end{tabular}\label{table:t5}
\end{table*}

\subsection{Ablation on Adaptive Merging }\label{apen:adaptive}

We find that when applying the merging method of (IA)$^3$ to LoRA, the performance is reduced much. Specifically, on XNLI the performance gets $39.5\%$ absolute reduction, while for XQuAD the reduction is $45.9\%$ absolute value. When applying the merging method of LoRA to (IA)$^3$, the performance also decreases compared to that of the adaptive merging method. For XNLI the reduction is $2.4\%$, while for XQuAD the reduction is $0.7\%$. The reduction is smaller than that for LoRA. This can be attributed to the fact that the fine-tuning of (IA)$^3$ is not as effective as that of LoRA and has a relatively minor impact on the overall model performance. 

\begin{table*}[hbpt]
\caption{Ablation study on adaptive merging method. \texttt{AdaMergeX} (adaptive) represents \texttt{AdaMergeX} with adaptive merging methods, while \texttt{AdaMergeX} (cross) represents \texttt{AdaMergeX} with cross merging methods, i.e., LoRA with merging method of (IA)$^3$ and vice versa. Increase \textcolor[RGB]{84,123,71}{$\uparrow$} and decrease \textcolor[RGB]{236,89,69}{$\downarrow$} are both compared to the baseline method Eng-Tune.}
  \centering
\footnotesize
  \scalebox{1.0}{
  \begin{tabular}{lll|lll}
    \toprule
 \multirow{1}*{\textbf{{Adapters}}} & \multirow{1}*{\textbf{{Tasks}}} & \multirow{1}*{\textbf{{Method}}} & es & vi & Avg.  \\
    \midrule
\multirow{6}*{LoRA} & \multirow{3}*{\begin{tabular}[c]{@{}l@{}}{{XNLI}}
\end{tabular}} & Eng-Tune & 60.5 & 47.0 & 53.8 \\
 &  & {\texttt{AdaMergeX} (adaptive)}  & \textbf{62.8} \textcolor[RGB]{84,123,71}{$\uparrow2.3$} & \textbf{49.2} \textcolor[RGB]{84,123,71}{$\uparrow2.2$} & \textbf{56.0} \textcolor[RGB]{84,123,71}{$\uparrow2.2$} \\  
& & {\texttt{AdaMergeX} (cross)}  & 17.6 \textcolor[RGB]{236,89,69}{$\downarrow42.9$} & 15.4 \textcolor[RGB]{236,89,69}{$\downarrow31.6$} & 16.5 \textcolor[RGB]{236,89,69}{$\downarrow37.3$} \\  \cmidrule{2-6}
 & \multirow{3}*{\begin{tabular}[c]{@{}l@{}}{{XQUAD}}
\end{tabular}} & Eng-Tune & 48.2 &  40.9 & 44.6 \\
   & & {\texttt{AdaMergeX} (adaptive)}  & \textbf{50.0} \textcolor[RGB]{84,123,71}{$\uparrow1.8$} & \textbf{41.7} \textcolor[RGB]{84,123,71}{$\uparrow0.8$} & \textbf{45.9} \textcolor[RGB]{84,123,71}{$\uparrow1.3$} \\  
& & {\texttt{AdaMergeX} (cross)}   & 0.0 \textcolor[RGB]{236,89,69}{$\downarrow48.2$} & 0.0 \textcolor[RGB]{236,89,69}{$\downarrow40.9$} & 0.0 \textcolor[RGB]{236,89,69}{$\downarrow44.6$} \\ \midrule
 \multirow{6}*{(IA)$^3$} & \multirow{3}*{\begin{tabular}[c]{@{}l@{}}{{XNLI}}
\end{tabular}} & Eng-Tune & 50.7 & 51.0 & 50.9 \\
  & & {\texttt{AdaMergeX} (adaptive)}  & \textbf{54.3} \textcolor[RGB]{84,123,71}{$\uparrow3.6$} & \textbf{58.8} \textcolor[RGB]{84,123,71}{$\uparrow7.8$} & \textbf{56.4} \textcolor[RGB]{84,123,71}{$\uparrow5.5$} \\  
& & {\texttt{AdaMergeX} (cross)}  & 50.9 \textcolor[RGB]{84,123,71}{$\uparrow0.2$} & 57.4 \textcolor[RGB]{84,123,71}{$\uparrow6.4$} & 54.2 \textcolor[RGB]{84,123,71}{$\uparrow3.1$} \\    \cmidrule{2-6}
& \multirow{3}*{\begin{tabular}[c]{@{}l@{}}{{XQUAD}}
\end{tabular}} & Eng-Tune & 47.6 & 35.1 & 41.4 \\
 &  & {\texttt{AdaMergeX} (adaptive)}  & \textbf{48.2} \textcolor[RGB]{84,123,71}{$\uparrow0.6$} & \textbf{35.7} \textcolor[RGB]{84,123,71}{$\uparrow0.6$} & \textbf{42.0} \textcolor[RGB]{84,123,71}{$\uparrow0.6$} \\  
& & {\texttt{AdaMergeX} (cross)}  & 47.5 \textcolor[RGB]{236,89,69}{$\downarrow0.1$} & 34.9 \textcolor[RGB]{236,89,69}{$\downarrow0.2$} & 41.3 \textcolor[RGB]{236,89,69}{$\downarrow0.1$}\\ 
    \bottomrule
  \end{tabular}}
  \label{table:abla_mer}
\end{table*}

\subsection{Ablation on Merging Modules}\label{apen:qv}

We present ablation on merging methods in Table \ref{table:abl_qv} and Table \ref{table:abla_all}.

\begin{table*}[hbpt]
  \centering
\footnotesize
  \scalebox{1.0}{
  \begin{tabular}{ll|lllllll}
    \toprule
  \multirow{1}*{\textbf{{Models}}} & \multirow{1}*{\textbf{{Method}}} & de & ru & fr & es & th & sw & Avg.  \\
    \midrule
 \multirow{3}*{\begin{tabular}[c]{@{}l@{}}{{XNLI}}
\end{tabular}} & Eng-Tune & 63.3 & 56.4 & 56.6 & 58.6 & 4.1 & 41.5 & 46.8 \\
   & {\texttt{AdaMergeX}}  & 63.8 & 57.2 & 58.2 & 58.9 & 3.7 & 41.8 & \textbf{47.3}\textcolor[RGB]{84,123,71}{$\uparrow0.5$} \\\midrule
 \multirow{3}*{\begin{tabular}[c]{@{}l@{}}{{XQuAD}}
\end{tabular}}
   & Eng-Tune & 9.8 & 8.7 & $-$ & 15.2  & 4.4 & $-$ & 9.5 \\
   & {\texttt{AdaMergeX}}  & 10.4 & 7.8 & $-$ & 21.4  & 5.4 & $-$ & \textbf{11.2}\textcolor[RGB]{84,123,71}{$\uparrow1.7$} \\
    \bottomrule
  \end{tabular}}
\caption{Llama2-7b on LoRA with fine-tuning target modules as $W^Q$, $W^V$ and merging target modules as $W^Q$, $W^V$.}
  \label{table:abl_qv}
\end{table*}

\begin{table*}[hbpt]
  \centering
\footnotesize
  \scalebox{1.0}{
  \begin{tabular}{ll|lllllll}
    \toprule
  \multirow{1}*{\textbf{{Models}}} & \multirow{1}*{\textbf{{Method}}} & de & ru & fr & es & th & sw & Avg.  \\
    \midrule
 \multirow{3}*{\begin{tabular}[c]{@{}l@{}}{{XNLI}}
\end{tabular}} & Eng-Tune & 54.0 & 54.0 & 58.2 & 60.5 & 3.3 & 31.8 & 43.6 \\
   & {\texttt{AdaMergeX}}  & 53.7 & 55.6 & 60.5 & 62.7 & 4.9 & 33.6 & \textbf{45.2}\textcolor[RGB]{84,123,71}{$\uparrow1.6$}\\\midrule
 \multirow{3}*{\begin{tabular}[c]{@{}l@{}}{{XQuAD}}
\end{tabular}}
   & Eng-Tune & 49.0 & 34.1  & $-$ & 48.2  & 31.0 & $-$ & 40.6 \\
   & {\texttt{AdaMergeX}}  & 50.2 & 32.9 & $-$ & 48.9  & 31.3 & $-$ & \textbf{40.8} \textcolor[RGB]{84,123,71}{$\uparrow0.2$} \\
    \bottomrule
  \end{tabular}}
\caption{Llama2-7b on LoRA with fine-tuning target modules as $W^Q$, $W^K$, $W^V$, $W^O$, $W_1$, $W_2$ and merging target modules as $W^Q$, $W^K$, $W^V$, $W^O$, $W_1$, $W_2$.}
  \label{table:abla_all}
\end{table*}

\end{document}